\documentclass[10pt, a4paper]{article}

\usepackage[final]{lrec2026} 

\usepackage{multirow}
\usepackage{adjustbox}
\usepackage{booktabs}
\usepackage{arydshln}
\usepackage{todonotes}
\usepackage{pgfplots}
\usepackage{caption}
\usepackage[framemethod=TikZ]{mdframed}
\usepackage{tikz}
\usepackage{color,xcolor,colortbl}
\usepackage{pgfplots}
\usepgfplotslibrary{fillbetween}
\usetikzlibrary{patterns}
\usepackage{subcaption}

\pgfplotsset{compat=1.18}

\newcommand{\quotes}[1]{``#1''}
\renewcommand{\vec}[1]{\mathbf{#1}}

\definecolor{lightgray}{rgb}{0.97,0.97,0.97} 
\definecolor{darkgray}{rgb}{0.8,0.8,0.8} 
\definecolor{lightgreenprompt}{HTML}{90EE90}

\definecolor{lightblue}{HTML}{5DA5DA}
\definecolor{lightorange}{HTML}{FAA43A}
\definecolor{lightgreen}{HTML}{60BD68}
\definecolor{lightpurple}{HTML}{B276B2}

\makeatletter
\def\adl@drawiv#1#2#3{%
        \hskip.5\tabcolsep
        \xleaders#3{#2.5\@tempdimb #1{1}#2.5\@tempdimb}%
                #2\z@ plus1fil minus1fil\relax
        \hskip.5\tabcolsep}
\newcommand{\cdashlinelr}[1]{%
  \noalign{\vskip\aboverulesep
           \global\let\@dashdrawstore\adl@draw
           \global\let\adl@draw\adl@drawiv}
  \cdashline{#1}
  \noalign{\global\let\adl@draw\@dashdrawstore
           \vskip\belowrulesep}}
\makeatother

\title{Extending Czech Aspect-Based Sentiment Analysis with Opinion Terms: Dataset and LLM Benchmarks}

\name{Jakub \v{S}m\'{i}d\textsuperscript{*, $\dagger$},
 Pavel P\v{r}ib\'{a}\v{n}\textsuperscript{*},
Pavel Kr\'{a}l\textsuperscript{*, $\dagger$}} 

\address{\textsuperscript{*}Department of Computer Science and Engineering\\
         \textsuperscript{$\dagger$}NTIS -- New Technologies for the Information Society\\
  University of West Bohemia in Pilsen, Faculty of Applied Sciences \\
          Univerzitní 2732/8, 301 00 Pilsen, Czech Republic \\
         \texttt{\{jaksmid, pribanp, pkral\}@kiv.zcu.cz}\\
         \url{https://nlp.kiv.zcu.cz}}

\abstract{
This paper introduces a novel Czech dataset in the restaurant domain for aspect-based sentiment analysis (ABSA), enriched with annotations of opinion terms. The dataset supports three distinct ABSA tasks involving opinion terms, accommodating varying levels of complexity. Leveraging this dataset, we conduct extensive experiments using modern Transformer-based models, including large language models (LLMs), in monolingual, cross-lingual, and multilingual settings. To address cross-lingual challenges, we propose a translation and label alignment methodology leveraging LLMs, which yields consistent improvements. Our results highlight the strengths and limitations of state-of-the-art models, especially when handling the linguistic intricacies of low-resource languages like Czech. A detailed error analysis reveals key challenges, including the detection of subtle opinion terms and nuanced sentiment expressions. The dataset establishes a new benchmark for Czech ABSA, and our proposed translation–alignment approach offers a scalable solution for adapting ABSA resources to other low-resource languages. 
 \\ \newline  }

\begin{document}

\maketitleabstract

\section{Introduction}
Aspect-based sentiment analysis (ABSA) is a fine-grained sentiment analysis task that extracts detailed information about entities and aspects. ABSA involves four sentiment elements~\citep{absa, SMID2025103073}: aspect term ($a$), aspect category ($c$), sentiment polarity ($p$), and opinion term ($o$). For example, in \textit{\quotes{Delicious tea}}, these correspond to \textit{\quotes{tea}}, \textit{\quotes{drinks quality}}, \textit{\quotes{positive}}, and \textit{\quotes{Delicious}}, respectively. Aspect and opinion terms may also be implicit, commonly annotated as \textit{\quotes{NULL}}; for instance, in \textit{\quotes{Tasty}}, the aspect term is implicit.

\begin{table}[ht!]
    \centering
    \begin{adjustbox}{width=\linewidth}
        \begin{tabular}{lll}
            \toprule
            \textbf{Task} & \textbf{Output} & \textbf{Example output}                                                                                                                 \\ \midrule
            ASTE & \{($a$, $o$, $p$)\} & \{(\quotes{tea}, \quotes{delicious}, POS)\} \\
            ASQP & \{($a$, $c$, $o$, $p$)\} & \{(\quotes{tea}, drinks, \quotes{delicious}, POS)\} \\
            ACOS & \{($a$, $c$, $o$, $p$)\} & \{(\quotes{tea}, drinks, \quotes{delicious}, POS), (\quotes{soup}, food, \quotes{NULL}, NEG)\}
 \\ \bottomrule
            \end{tabular}
    \end{adjustbox}
    \caption{Output format for selected ABSA tasks that involve opinion terms for the input sentence: \textit{\quotes{The tea was delicious, unlike soup}}.}
	\label{tab:absa-tasks}
\end{table}

Research has gradually evolved from simple ABSA tasks (e.g. aspect term extraction) to compound tasks requiring linked predictions of multiple sentiment elements. More recently, opinion terms have gained increasing attention, with tasks such as aspect sentiment triplet extraction (ASTE)~\citep{aste}, aspect sentiment quad prediction (ASQP)~\citep{zhang-etal-2021-aspect-sentiment}, and aspect category opinion sentiment prediction (ACOS)~\citep{cai-etal-2021-aspect}. Table~\ref{tab:absa-tasks} provides input and output examples for such tasks. Modern approaches often cast these tasks as text generation problems using pre-trained sequence-to-sequence models~\citep{zhang-etal-2021-towards-generative, gou-etal-2023-mvp}. 

Large language models (LLMs) like LLaMA~3.1~\citep{dubey2024llama3herdmodels} have advanced natural language processing substantially. While smaller fine-tuned Transformer-based models still outperform LLMs on ABSA when sufficient training data is available~\citep{gou-etal-2023-mvp, zhang-etal-2024-sentiment}, recent studies highlight the potential of fine-tuned LLMs for ABSA~\citep{smid-etal-2024-llama, smid2025largelanguagemodelsczech}. Moreover, parameter-efficient methods such as QLoRA~\citep{qlora} make LLM fine-tuning feasible on limited hardware.

Over time, several datasets have been developed for ABSA, including SemEval-2014--2016~\citep{pontiki-etal-2014-semeval, pontiki-etal-2015-semeval, pontiki-etal-2016-semeval} and SentiHood~\citep{saeidi-etal-2016-sentihood}. Most focus on English, with SemEval-2016 also covering several other languages. Later extensions introduced opinion term annotations, enabling tasks such as ASTE~\citep{aste, xu-etal-2020-position}, ASQP~\citep{zhang-etal-2021-aspect-sentiment}, and ACOS~\citep{cai-etal-2021-aspect}. For Czech, existing ABSA datasets~\citep{steinberger-etal-2014-aspect, hercig2016unsupervised, tamchyna2015czech, smid-etal-2024-czech} support simple or compound tasks but lack opinion term annotations, restricting research on more complex setups. Since the datasets labelled with all four sentiment elements are only available in English, it is not possible to perform cross-lingual comparisons with other languages.

To address this gap, we introduce a new Czech dataset with opinion term annotations, supporting three compound tasks (ASTE, ASQP, ACOS). To our knowledge, it is the first dataset beyond English to allow quadruplet-level tasks such as ASQP and ACOS. The dataset and code is publicly released to foster further research\footnote{Code and dataset are available at the anonymous repository: \url{https://github.com/biba10/Czech-ABSA-Opinion-Dataset-Benchmark}}. We benchmark modern Transformer-based models and LLMs in monolingual, multilingual, and cross-lingual settings. To the best of our knowledge, we are the first to explore cross-lingual configurations for all three tasks. Through error analysis, we further highlight the challenges posed by complex ABSA in Czech.

Our main contributions are as follows: 1) We present a new Czech dataset tailored for compound ABSA tasks, complete with opinion term annotations. 2) We evaluate leading large language models in zero-shot, few-shot, and fine-tuning scenarios, analysing their strengths and limitations. 3) We compare fine-tuned LLMs with a multilingual sequence-to-sequence baseline. 4) We propose a novel method for cross-lingual transfer based on data translation and label alignment with LLMs. 5) We conduct an error analysis to highlight the main challenges posed by the dataset and future research directions.

\section{Related Work}
We review existing ABSA datasets with opinion term annotations, ABSA datasets for Czech, and prior work on ABSA methods, with an emphasis on Czech.

\subsection{ABSA Datasets}

The USAGE dataset~\citep{klinger-cimiano-2014-usage} provides English and German product reviews annotated with aspect and opinion terms. However, the annotations are not linked, which limits the dataset's suitability for tasks requiring aspect–opinion pairing.

Most English ABSA resources build on the SemEval 2014--2016 datasets~\citep{pontiki-etal-2014-semeval, pontiki-etal-2015-semeval, pontiki-etal-2016-semeval}, which focus on the restaurant domain and include aspect terms but no opinion terms. \citet{fan-etal-2019-target} added opinion terms and linked them to aspect terms, enabling aspect–opinion pair extraction, though sentiment polarity and aspect category were omitted. \citet{aste} extended this work by merging annotations, reintroducing polarity, and producing data suitable for ASTE. Both works exclude sentences with implicit aspects.

Quadruplet-level datasets further enriched the annotations. The ASQP dataset~\citep{zhang-etal-2021-aspect-sentiment} reintroduced implicit aspects and added aspect categories. The ACOS dataset~\citep{cai-etal-2021-aspect} went further by including implicit opinion terms (ASQP only has explicit ones) and expanding coverage to laptops as well as restaurants.

For Czech ABSA, resources remain scarce. The earliest dataset~\citep{steinberger-etal-2014-aspect} provides restaurant reviews in the SemEval-2014 format. \citet{tamchyna2015czech} created a dataset of IT product reviews with aspect and sentiment annotations. \citet{hercig2016unsupervised} expanded the Czech restaurant dataset, retaining the SemEval-2014 format but without linking aspects to categories. More recently, \citet{smid-etal-2024-czech} introduced a Czech dataset in the SemEval-2016 format with linked aspect terms, categories, and sentiment polarities, enabling compound tasks. However, no Czech ABSA dataset includes opinion term annotations.

\subsection{Aspect-Based Sentiment Analysis}
Early Czech ABSA work~\citep{steinberger-etal-2014-aspect, tamchyna2015czech, hercig2016unsupervised} relied on traditional machine learning methods such as maximum entropy classifiers, later replaced by neural networks~\citep{SLON-Lenc2016Neural}. More recent studies adopt Transformer-based models~\citep{vaswani2017attention}, including prompt-based approaches~\citep{smid-priban-2023-prompt}, multitask learning~\citep{priban-prazak-2023-improving}, and advanced fine-tuned architectures~\citep{smid-etal-2024-czech}. This mirrors developments in English ABSA, where recent work often frames the task as text generation using sequence-to-sequence models~\citep{zhang-etal-2021-towards-generative,zhang-etal-2021-aspect-sentiment,gao-etal-2022-lego,mao-etal-2022-seq2path,gou-etal-2023-mvp,xianlong-etal-2023-tagging}.

Large language models have also been explored. In zero- and few-shot settings, they typically underperform compared to fine-tuned baselines~\citep{gou-etal-2023-mvp, zhang-etal-2024-sentiment}, but fine-tuned LLMs have shown strong results in English~\citep{smid-etal-2024-llama}, Czech~\citep{smid2025largelanguagemodelsczech}, and other languages~\citep{icaart25, smid-etal-2025-laca, Wu_2025}.

\section{Dataset Construction}
We build upon the \texttt{CsRest-M} dataset~\citep{smid-etal-2024-czech}, which contains annotations for aspect terms, aspect categories, and sentiment polarity triplets, and is already divided into training, validation, and test sets. Our primary enhancement is the addition of opinion term annotations, which extend the dataset to support the ASTE, ASQP, and ACOS tasks.

\begin{figure*}[ht!]
    \centering
    \includegraphics[width=0.99\linewidth]{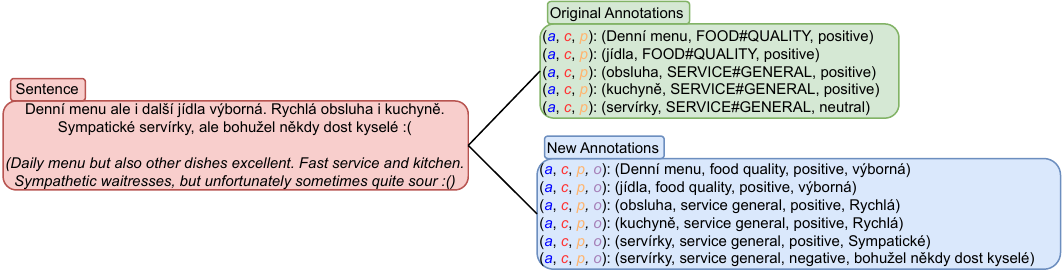}
    \caption{Example of the original annotations (top right) and the updated versions after our modifications (bottom right).}
    \label{fig:annotations}
\end{figure*}

\subsection{Annotation Process}
\label{sec:annotation}

Before starting, the annotators developed detailed guidelines, drawing inspiration from the USAGE dataset~\citep{klinger-cimiano-2014-usage} as well as the English ACOS~\citep{cai-etal-2021-aspect} and ASQP~\citep{zhang-etal-2021-aspect-sentiment} datasets. The final version of these guidelines is described in the following section, where we summarize the key principles. The development process was iterative: after annotating a few hundred samples, the annotators reviewed their work, discussed any ambiguities, and refined the guidelines accordingly. This approach promoted high inter-annotator agreement and resolved most issues early in the process.

In most cases, the original triplets were retained, with annotators tasked with adding corresponding opinion terms. However, certain sentences required the addition of new quadruplets or adjustments to sentiment polarity. For instance, the phrase \textit{\quotes{Velmi rychlá a milá obsluha}} (\textit{\quotes{Very fast and friendly service}}) was originally annotated with a single triplet (\textit{\quotes{obsluha}}, \textit{\quotes{service general}}, \textit{\quotes{positive}}) for the aspect term \textit{\quotes{obsluha}} (\textit{\quotes{service}}). In our dataset, this was expanded to two quadruplets: one linking the opinion term \textit{\quotes{Velmi rychlá}} (\textit{\quotes{Very fast}}) and another with \textit{\quotes{milá}} (\textit{\quotes{friendly}}). Similarly, for the phrase \textit{\quotes{Sympatické servírky, ale bohužel někdy dost kyselé}} (\textit{\quotes{Sympathetic waitresses, but unfortunately sometimes quite sour}}), the original annotation marked the sentiment as \textit{\quotes{neutral}} due to its mixed nature for the aspect term \textit{\quotes{servírky}} (\textit{\quotes{waitresses}}). We refined this by creating two quadruplets: one with \textit{\quotes{positive}} polarity for the opinion term \textit{\quotes{Sympatické}} (\textit{\quotes{Sympathetic}}), and another with \textit{\quotes{negative}} polarity for the opinion term \textit{\quotes{bohužel někdy dost kyselé}} (\textit{\quotes{unfortunately sometimes quite sour}}).

A key decision was whether to include modifiers for opinion terms, such as \textit{\quotes{velmi}} (\textit{\quotes{very}}) in \textit{\quotes{velmi rychlá}} (\textit{\quotes{very fast}}). Existing datasets vary: the USAGE dataset~\citep{klinger-cimiano-2014-usage} includes modifiers, while ASTE and ACOS datasets~\citep{fan-etal-2019-target, aste, cai-etal-2021-aspect} do not. The ASQP dataset~\citep{zhang-etal-2021-aspect-sentiment} is inconsistent, sometimes including modifiers and sometimes not. We chose to annotate modifiers for two reasons: (1) in Czech, modifiers can significantly affect sentiment intensity, distinguishing mildly positive/negative from strongly positive/negative sentiment, with the former annotated as neutral in related work~\citep{pontiki-etal-2015-semeval, pontiki-etal-2016-semeval, smid-etal-2024-czech}; and (2) modifiers may support future extensions of sentiment polarity annotations, such as introducing \textit{\quotes{very positive}} or \textit{\quotes{very negative}} categories. This decision may complicate cross-lingual experiments with English ASTE, ASQP, and ACOS datasets, where modifiers are generally omitted. We also annotated implicit opinion terms (marked as \textit{\quotes{NULL}}) to make the dataset suitable for the ACOS task. 

The main annotation tasks for each review sentence were as follows: 
\begin{quote} 
\textbf{Identify opinion terms for each triplet:} Assign an opinion term to each annotated triplet. If no explicit opinion term exists, use \textit{\quotes{NULL}}. Introduce additional quadruplets if multiple opinion terms exist for a single aspect term. Adjust sentiment polarity if conflicting opinions are expressed.
\end{quote}

After completing the annotation process, we applied two further modifications to ensure consistency with related datasets. First, the aspect categories in the original dataset were provided in the \textit{ENTITY\#ATTRIBUTE} format, as in the SemEval datasets. To align with English datasets for ACOS, ASQP, and ASTE, we replaced the \quotes{\#} symbol with a space and converted the categories to lowercase (e.g. \textit{\quotes{FOOD\#QUALITY}} became \textit{\quotes{food quality}}). Second, we normalized the text to improve readability and processing, including reducing multiple consecutive spaces to a single space and inserting spaces to separate punctuation from words, following conventions used in English datasets. Figure~\ref{fig:annotations} illustrates the transition from the original annotations to the updated versions.

\begin{table*}[ht!]
\centering
\begin{adjustbox}{width=0.95\linewidth}
\begin{tabular}{@{}rlrrrrrrrrrrrrrrrr@{}}
\toprule
                       & \multirow{2}{*}{\textbf{Category}} & \multicolumn{4}{c}{\textbf{Train}} & \multicolumn{4}{c}{\textbf{Dev}} & \multicolumn{4}{c}{\textbf{Test}} & \multicolumn{4}{c}{\textbf{Total}} \\ \cmidrule(lr){3-6}  \cmidrule(lr){7-10} \cmidrule(lr){11-14}  \cmidrule(lr){15-18} 
                       &                                    & Pos    & Neg    & Neu    & Tot     & Pos    & Neg    & Neu    & Tot   & Pos    & Neg    & Neu    & Tot    & Pos     & Neg    & Neu    & Tot    \\ \midrule
\multirow{12}{*}{\rotatebox[origin=c]{90}{ACOS}} & ambience general & 349 & 97 & 30 & 476 & 40 & 12 & 3 & 55 & 121 & 46 & 8 & 175 & 510 & 155 & 41 & 706   \\
& ambience general & 349 & 97 & 30 & 476 & 40 & 12 & 3 & 55 & 121 & 46 & 8 & 175 & 510 & 155 & 41 & 706   \\
& drinks prices & 9 & 14 & 5 & 28 & 1 & 1 & 4 & 6 & 4 & 10 & 0 & 14 & 14 & 25 & 9 & 48   \\
& drinks quality & 191 & 50 & 20 & 261 & 24 & 5 & 2 & 31 & 66 & 14 & 4 & 84 & 281 & 69 & 26 & 376   \\
& drinks style\_options & 33 & 21 & 7 & 61 & 3 & 2 & 0 & 5 & 19 & 4 & 0 & 23 & 55 & 27 & 7 & 89   \\
& food prices & 41 & 56 & 18 & 115 & 7 & 5 & 2 & 14 & 9 & 21 & 13 & 43 & 57 & 82 & 33 & 172   \\
& food quality & 942 & 400 & 108 & 1,450 & 89 & 46 & 7 & 142 & 341 & 117 & 37 & 495 & 1,372 & 563 & 152 & 2,087   \\
& food style\_options & 137 & 118 & 23 & 278 & 15 & 8 & 6 & 29 & 39 & 53 & 4 & 96 & 191 & 179 & 33 & 403   \\
& location general & 30 & 1 & 0 & 31 & 1 & 3 & 0 & 4 & 18 & 3 & 0 & 21 & 49 & 7 & 0 & 56   \\
& restaurant general & 544 & 241 & 30 & 814 & 59 & 38 & 5 & 102 & 198 & 101 & 15 & 314 & 801 & 380 & 50 & 1,231   \\
& restaurant miscellaneous & 29 & 63 & 13 & 105 & 3 & 7 & 2 & 12 & 9 & 27 & 7 & 43 & 41 & 97 & 22 & 160   \\
& restaurant prices & 57 & 48 & 27 & 132 & 3 & 3 & 2 & 8 & 25 & 14 & 9 & 48 & 85 & 65 & 38 & 188   \\
& service general & 607 & 326 & 53 & 986 & 61 & 46 & 8 & 115 & 217 & 132 & 18 & 367 & 885 & 504 & 79 & 1,468   \\ \cdashlinelr{2-18}
& Total & 2,969 & 1,435 & 334 & 4,738 & 306 & 176 & 41 & 523 & 1,066 & 542 & 115 & 1,723 & 4,341 & 2,153 & 490 & 6,984   \\ \midrule
\multirow{12}{*}{\rotatebox[origin=c]{90}{ASQP}} & ambience general & 326 & 65 & 19 & 410 & 38 & 12 & 2 & 52 & 114 & 37 & 8 & 159 & 478 & 114 & 29 & 621   \\
& ambience general & 326 & 65 & 19 & 410 & 38 & 12 & 2 & 52 & 114 & 37 & 8 & 159 & 478 & 114 & 29 & 621   \\
& drinks prices & 6 & 8 & 2 & 16 & 1 & 1 & 4 & 6 & 4 & 7 & 0 & 11 & 11 & 16 & 6 & 33   \\
& drinks quality & 167 & 33 & 17 & 217 & 23 & 5 & 2 & 30 & 61 & 9 & 3 & 73 & 251 & 47 & 22 & 320   \\
& drinks style\_options & 27 & 11 & 4 & 42 & 3 & 2 & 0 & 5 & 16 & 3 & 0 & 19 & 46 & 16 & 4 & 66   \\
& food prices & 33 & 13 & 14 & 60 & 7 & 3 & 2 & 12 & 8 & 11 & 11 & 30 & 48 & 27 & 27 & 102   \\
& food quality & 865 & 260 & 86 & 1,211 & 83 & 36 & 7 & 126 & 327 & 85 & 35 & 447 & 1,275 & 381 & 128 & 1,784   \\
& food style\_options & 106 & 51 & 12 & 169 & 11 & 7 & 3 & 21 & 34 & 31 & 1 & 66 & 151 & 89 & 16 & 256   \\
& location general & 24 & 0 & 0 & 24 & 0 & 3 & 0 & 3 & 12 & 2 & 0 & 14 & 36 & 5 & 0 & 41   \\
& restaurant general & 439 & 141 & 14 & 594 & 48 & 25 & 2 & 75 & 170 & 57 & 10 & 237 & 657 & 223 & 26 & 906   \\
& restaurant miscellaneous & 19 & 10 & 4 & 33 & 1 & 2 & 0 & 3 & 5 & 4 & 0 & 9 & 25 & 16 & 4 & 45   \\
& restaurant prices & 49 & 28 & 16 & 93 & 2 & 2 & 2 & 6 & 23 & 8 & 5 & 36 & 74 & 38 & 23 & 135   \\
& service general & 558 & 151 & 28 & 737 & 58 & 27 & 5 & 90 & 206 & 72 & 11 & 289 & 822 & 250 & 44 & 1,116   \\ \cdashlinelr{2-18}
& Total & 2,619 & 771 & 216 & 3,606 & 275 & 125 & 29 & 429 & 980 & 326 & 84 & 1,390 & 3,874 & 1,222 & 329 & 5,425  \\ \midrule 
\rotatebox[origin=c]{90}{ASTE} & Total & 2,169 & 603 & 178 & 2,950 & 234 & 99 & 25 & 358 & 808 & 258 & 64 & 1,130 & 3,211 & 960 & 267 & 4,438 \\
                       \bottomrule
\end{tabular}
\end{adjustbox}
\caption{Detailed statistics of our datasets by aspect category and sentiment polarity. Columns Pos, Neg, Neu, and Tot denote counts for positive, negative, neutral, and total instances, respectively.}
    \label{tab:detailed_stats}
\end{table*}

Two native Czech speakers with prior experience in ABSA annotation performed the annotation. A third annotator, also experienced in ABSA, assisted in reviewing and resolving disagreements, supporting the other two annotators to ensure consistency and high quality. The final ACOS dataset contains 3{,}000 sentences with almost 7{,}000 annotated quadruplets. No instances of content that could be considered discriminatory or overtly racist were found in the dataset, though some reviews contain mild offensive language typical of user-generated reviews.

\subsection{Derived ASTE and ASQP Datasets}
From the ACOS dataset, we derived task-specific variants. For ASQP, we removed quadruplets with implicit opinion terms (\textit{\quotes{NULL}}) and filtered out sentences containing no remaining quadruplets after this exclusion. For ASTE, we omitted aspect category annotations, excluded triplets with implicit aspect or opinion terms, merged identical triplets, and removed sentences without any remaining triplets after these steps.

\subsection{Dataset Statistics}
Table~\ref{tab:detailed_stats} presents a detailed distribution of our datasets, including sentiment polarities and, for ASQP and ACOS, aspect categories. The most frequent polarity is \textit{\quotes{positive}}, while \textit{\quotes{neutral}} is relatively rare, accounting for only about 6\% of the tuples. The three most common aspect categories are \textit{\quotes{food quality}}, \textit{\quotes{restaurant general}}, and \textit{\quotes{service general}}, whereas the least frequent are \textit{\quotes{location general}}, \textit{\quotes{drinks prices}}, and \textit{\quotes{drinks style\_options}}.

\begin{table}[ht!]
\centering
\begin{adjustbox}{width=0.85\linewidth}
\begin{tabular}{@{}lllrrrrr@{}}
\toprule
\textbf{Dataset}                                                                                             & \textbf{Lang}       & \textbf{Split} & \textbf{Sentences} & \textbf{Tuples} & \textbf{IA}    & \textbf{IO}    & \textbf{IA \& IO} \\ \midrule
\multirow{8}{*}{\rotatebox[origin=c]{90}{\begin{tabular}[c]{@{}c@{}}ACOS\\ En: \citep{cai-etal-2021-aspect}\end{tabular}}}             & \multirow{4}{*}{\begin{tabular}[c]{@{}l@{}}Cs\\ (Ours)\end{tabular}} & Train          & 2,018     & 4,738  & 1,038 & 1,132 & 389      \\
                                                                                                             &                     & Dev            & 230       & 523    & 104   & 94    & 34       \\
                                                                                                             &                     & Test           & 752       & 1,723  & 367   & 333   & 113      \\
                                                                                                             &                     & Total          & 3,000     & 6,984  & 1,509 & 1,559 & 536      \\ \cdashlinelr{2-8}
                                                                                                             & \multirow{4}{*}{En} & Train          & 1,530     & 2,484  & 607   & 448   & 233      \\
                                                                                                             &                     & Dev            & 171       & 261    & 60    & 54    & 26       \\
                                                                                                             &                     & Test           & 583       & 916    & 213   & 198   & 91       \\
                                                                                                             &                     & Total          & 2,283     & 3,660  & 880   & 700   & 350      \\ \midrule
\multirow{8}{*}{\rotatebox[origin=c]{90}{\begin{tabular}[c]{@{}c@{}}ASQP\\ En: \citep{zhang-etal-2021-aspect-sentiment}\end{tabular}}} & \multirow{4}{*}{\begin{tabular}[c]{@{}l@{}}Cs\\ (Ours)\end{tabular}} & Train          & 1,594     & 3,606  & 649   & 0     & 0        \\
                                                                                                             &                     & Dev               & 198       & 429    & 70    & 0     & 0        \\
                                                                                                             &                     & Test               & 616       & 1,390  & 254   & 0     & 0        \\
                                                                                                             &                     & Total               & 2,408     & 5,425  & 973   & 0     & 0        \\ \cdashlinelr{2-8}
                                                                                                             & \multirow{4}{*}{En} & Train          & 1,264     & 1,989  & 446   & 0     & 0        \\
                                                                                                             &                     & Dev            & 316       & 507    & 104   & 0     & 0        \\
                                                                                                             &                     & Test           & 544       & 799    & 179   & 0     & 0        \\
                                                                                                             &                     & Total          & 2,124     & 3,295  & 729   & 0     & 0        \\ \midrule
\multirow{8}{*}{\rotatebox[origin=c]{90}{\begin{tabular}[c]{@{}c@{}}ASTE\\ En: \citep{aste}\end{tabular}}}                             & \multirow{4}{*}{\begin{tabular}[c]{@{}l@{}}Cs\\ (Ours)\end{tabular}} & Train          & 1,321     & 2,950  & 0     & 0     & 0        \\
                                                                                                             &                     & Dev            & 161       & 358    & 0     & 0     & 0        \\
                                                                                                             &                     & Test           & 505       & 1,130  & 0     & 0     & 0        \\
                                                                                                             &                     & Total          & 1,987     & 4,438  & 0     & 0     & 0        \\ \cdashlinelr{2-8}
                                                                                                             & \multirow{4}{*}{En} & Train          & 857       & 1,394  & 0     & 0     & 0        \\
                                                                                                             &                     & Dev            & 210       & 339    & 0     & 0     & 0        \\
                                                                                                             &                     & Test           & 326       & 514    & 0     & 0     & 0        \\
                                                                                                             &                     & Total          & 1,393     & 2,247  & 0     & 0     & 0        \\ \bottomrule
\end{tabular}
\end{adjustbox}
\caption{Comparison of our datasets with English counterparts from the SemEval-2016 restaurant domain~\citep{pontiki-etal-2016-semeval}. IA = implicit aspect terms, IO = implicit opinion terms, IA \& IO = tuples containing both an implicit aspect and an implicit opinion term.}
\label{tab:data_comparison}
\end{table}

Table~\ref{tab:data_comparison} compares our datasets with existing English restaurant-domain datasets. Our datasets are substantially larger, both in the number of sentences and, more importantly, in the number of annotated tuples. For instance, our ASTE dataset contains nearly twice as many tuples as its English counterpart (4,438 vs 2,247). The distribution of implicit aspect and opinion terms in our ACOS and ASQP datasets is similar to that in English datasets, differing by only a few percentage points.

\subsection{Inter-annotator Agreement}

Following prior work~\citep{steinberger-etal-2014-aspect, pontiki-etal-2016-semeval, hercig2016unsupervised, cai-etal-2021-aspect, smid-etal-2024-czech}, we measure inter-annotator agreement (IAA) using strict quadruplet-matching F1, treating one annotator’s labels as gold and the other’s as predictions. This metric is standard in ABSA, where annotations consist of structured, multi-label quadruplets with variable cardinality per instance, making chance-corrected coefficients such as Cohen’s kappa less appropriate. After the first 100 examples, IAA was 63\%. Following guideline refinement and discussion of annotation issues, it rose to 76\% on the subsequent 100 examples. The final IAA over the entire dataset is 85\%, indicating substantial agreement and remaining comparable to previously reported results for English datasets~\citep{cai-etal-2021-aspect}.

Most disagreements stemmed from opinion term annotations, which were the main addition to the original dataset. The primary challenge concerned implicit opinion terms, where annotators sometimes disagreed on whether to treat an expression as explicit (e.g. annotating \textit{\quotes{doporučuji}} (\textit{\quotes{recommended}})), or leave it implicit. Other disagreements involved partially overlapping opinion terms, where one annotator included more words than the other, and modifiers that were occasionally omitted during the early stages. Most of these, along with additional early-phase inconsistencies, such as failing to create separate quadruplets for multiple opinion terms, were mitigated through iterative refinement of the guidelines (see Section~\ref{sec:annotation} for details).

\section{Experiments \& Setup}
We evaluate our new Czech ABSA dataset on three tasks: ASTE, ACOS, and ASQP. The primary evaluation metric is micro F1-score. For fine-tuning experiments, results are averaged over five runs with different random seeds, whereas zero-shot and few-shot experiments are performed with a single run, as they do not involve stochastic parameter updates. A predicted tuple is considered correct only if all of its components exactly match the corresponding gold tuple.

\subsection{Sequence-to-Sequence Models}
Following prior work on Czech ABSA~\citep{smid-priban-2023-prompt, smid-etal-2024-czech}, we fine-tune large mT5~\citep{xue-etal-2021-mt5}, Transformer-based encoder–decoder (sequence-to-sequence) model. The encoder produces contextualized representations $\vec{e}$ of the input text. The decoder estimates $P_{\vec{\Theta}}(y|\vec{e})$ over the output sequence $y$, generating each token $y_i$ conditioned on $\vec{e}$ and the previously generated tokens before the $i$-th step, denoted as $y_{<i}$.

\begin{figure}[ht!]
    \centering
    \includegraphics[width=\linewidth]{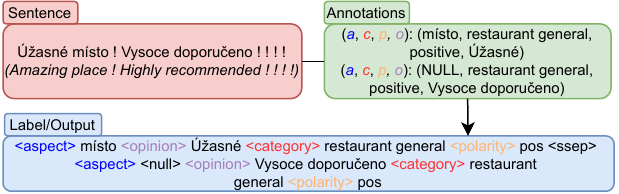}
    \caption{Example of converting ABSA annotations into output sequences for sequence-to-sequence models. Special tokens represent aspect terms, opinion terms, categories, sentiment polarities, and tuple separators.}
    \label{fig:seq2seq}
\end{figure}

We transform ABSA labels into a textual format using six special tokens: \texttt{<aspect>} for aspect terms, \texttt{<opinion>} for opinion terms, \texttt{<category>} for aspect categories, \texttt{<polarity>} for sentiment polarities, \texttt{<null>} for implicit terms, and \texttt{<ssep>} to separate tuples. We abbreviate sentiment polarity to its first three letters (e.g. \textit{\quotes{pos}} for \textit{\quotes{positive}}), and order the sentiment elements as $a \to o \to c \to p$~\citep{gou-etal-2023-mvp}. Figure~\ref{fig:seq2seq} illustrates the conversion process for a given input. The ASTE task does not require the \texttt{<null>} and \texttt{<category>} tokens.

During fine-tuning, we update all model parameters and minimize the negative log-likelihood as
\begin{equation}
    \mathcal{L} = -\sum_{i=1}^n\log p_{{\Theta}}(y_i|\vec{e},y_{<i}),
\end{equation}
where $n$ is the length of the target sequence $y$.

\subsection{Large Language Models}
We evaluate decoder-only LLMs in zero-shot, few-shot, and fine-tuning settings. Unlike encoder–decoder models, decoder-only models generate tokens autoregressively without a dedicated encoder. Their training loss follows the same principle as sequence-to-sequence models, but conditions only on previously generated tokens.

\begin{figure}[ht!]
    \centering
    \includegraphics[width=\linewidth]{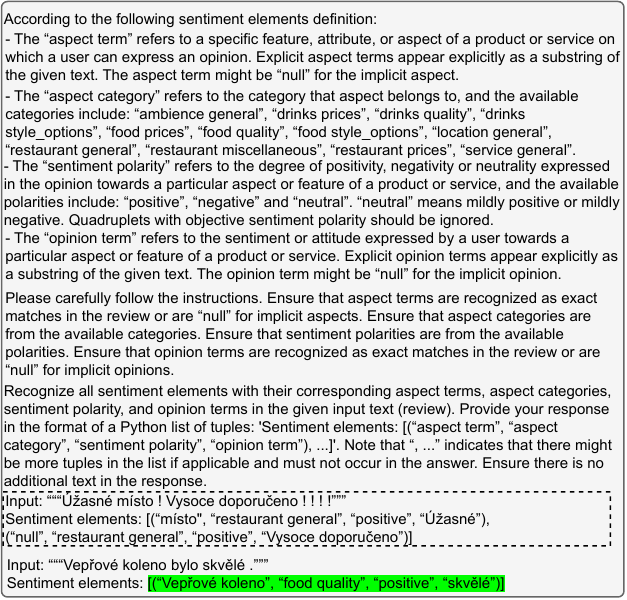}
    \caption{Prompt for the ACOS task with example input, expected output (green box), and one demonstration (dashed box, used only in few-shot scenarios).}
    \label{fig:prompt}
\end{figure}

We adopt prompts from prior Czech ABSA work~\citep{smid-etal-2024-llama,smid2025largelanguagemodelsczech}. Few-shot prompts use the first ten training examples, which provide sufficient coverage for fair comparison~\citep{smid2025largelanguagemodelsczech}. These examples are representative and unsorted, ensuring diversity beyond sentiment cues. Figure~\ref{fig:prompt} shows an ACOS prompt with one demonstration and its expected output. We adapt the same format for ASQP (without implicit opinions) and ASTE (excluding aspect categories and implicit terms). Prompts are written in English, as prompt language has little impact on Czech ABSA performance~\citep{smid2025largelanguagemodelsczech}.

We fine-tune LLMs with QLoRA~\citep{qlora}, which applies 4-bit quantization to a frozen backbone and learns only a small set of LoRA weights~\citep{hu2022lora}, substantially reducing memory requirements.

We evaluate a range of models for zero-shot and few-shot experiments: GPT-4o mini~\citep{openai2024gpt4o}, Orca~2 (13B)~\citep{mitra2023orca}, LLaMA~3.1 (8B, 70B), LLaMA~3.3 (70B)~\citep{dubey2024llama3herdmodels}, Gemma~3 (4B, 12B, 27B)~\citep{gemmateam2025gemma3technicalreport}, and Aya~23 (8B, 35B)~\citep{aryabumi2024aya23openweight}. Fine-tuning is restricted to models up to 12B parameters and uses the prompt shown in Figure~\ref{fig:prompt}. All models are open-source except GPT-4o mini.

\subsection{Multilingual and Cross-lingual Experiments}
Beyond monolingual setups, we perform multilingual and cross-lingual experiments. For English, we use datasets derived from the SemEval-2016 restaurant domain~\citep{pontiki-etal-2016-semeval}: ASTE~\citep{aste}, ACOS~\citep{cai-etal-2021-aspect}, and ASQP~\citep{zhang-etal-2021-aspect-sentiment}.

In multilingual experiments, models are trained on the union of Czech and English data and evaluated on Czech test sets, with model selection based on combined validation performance across both languages.

In cross-lingual settings, English serves as the source language and Czech as the target. Model selection is based on English validation data. We test two variants: (1) fine-tuning only on the English dataset, and (2) fine-tuning on English plus its machine-translated Czech counterpart.

\begin{figure}[ht!]
    \centering
    \includegraphics[width=\linewidth]{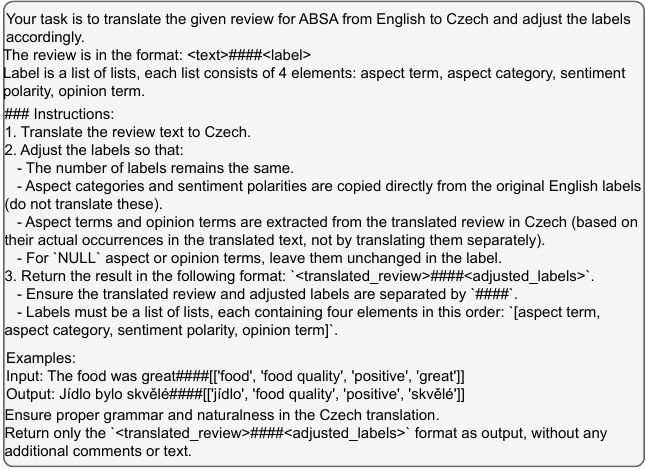}
    \caption{Prompt for translating the ACOS dataset from English to Czech with aligned labels. The full prompt contains five different representative input/output examples.}
    \label{fig:prompttr}
\end{figure}

To create aligned translations, we use GPT-4o mini with detailed instructions (Figure~\ref{fig:prompttr}). Because translation alters word counts and positions, we filter outputs by tuple counts, categories, sentiment polarities, and implicit/explicit markers. The primary source of errors -- explicit terms not matching the translated text -- occurred in about 10\% of cases. To our knowledge, this is the first use of LLMs for ABSA data translation with alignment. Compared to traditional methods such as FastAlign~\citep{dyer-etal-2013-simple}, which is often unreliable, or symbol-marking~\citep{zhang-etal-2021-cross}, which can drop or misalign labels, our approach provides a more robust alternative.

\subsection{Hyperparameters}

We employ all open-source models from the HuggingFace Transformers library\footnote{\url{https://github.com/huggingface/transformers}}~\citep{wolf-etal-2020-transformers}. All experiments use greedy search decoding and run on a single NVIDIA L40 GPU with 48 GB memory. Considering preprocessing, we lowercase all sentences and labels to maintain consistency across monolingual, multilingual, and cross-lingual experiments, preventing mismatches between datasets (some already lowercased) and mitigates performance degradation caused by case differences.

We fine-tune mT5 for 20 epochs with batch size 16, learning rate 1e-4, and the AdamW optimizer~\citep{loshchilov2017decoupled}, selected for consistent validation performance across tasks. 

For QLoRA fine-tuning, we use 4-bit NormalFloat (NF4) with double quantization and bf16 computation. Training parameters include batch size 16, constant learning rate $2e-4$, AdamW optimizer, and LoRA applied to all Transformer linear layers, with $r=64$ and $\alpha=16$. For Gemma~3, we follow prior recommendations~\citep{smid2025largelanguagemodelsczech} and use $r=64$ and $\alpha=128$. Training runs for up to 5 epochs, selecting the best model by validation loss. Loss is computed only on model-generated tokens, excluding prompts~\citep{mitra2023orca}. Zero-shot and few-shot models are quantized to 4 bits, as they perform similarly to full-precision models~\citep{qlora, smid-etal-2024-llama}.

\subsection{Results}

\begin{table}[ht!]
\centering
\begin{adjustbox}{width=0.85\linewidth}
\begin{tabular}{@{}llrrrr@{}}
\toprule
 \textbf{Type}                       & \textbf{Model}                        & \textbf{ASTE}  & \textbf{ASQP}        & \textbf{ACOS}  & \textbf{AVG}   \\ \midrule
 \multirow{10}{*}{\rotatebox[origin=c]{90}{Zero-shot}}         & Aya~23~8B                             & 25.59          & 8.76                 & 5.97           & 13.44          \\
                                &                                      Aya~23~35B                            & 37.34          & 26.00                & 23.18          & 28.84          \\
                                &                                      Gemma 3 4B                            & 30.87          & 15.20                & 8.23           & 18.10          \\
                                &                                      Gemma 3 12B                           & 51.64          & 31.75                & 28.00          & 37.13          \\
                                &                                      Gemma 3 27B                           & 53.27          & 41.27                & \textbf{35.43} & 43.33          \\
                                &                                      LLaMA~3.1~8B & 29.89          & 7.37                 & 3.50           & 13.59          \\
                                &                                      LLaMA~3.1~70B                         & 45.94          & 31.86                & 26.49          & 34.76          \\
                                &                                      LLaMA~3.3~70B                         & 50.21          & 38.29                & 32.08          & 40.19          \\
                                &                                      Orca~2~13B                            & 15.24          & 10.21                & 8.09           & 11.18          \\
                                &                                      GPT-4o mini                           & \textbf{56.43} & \textbf{42.12}       & 33.77          & \textbf{44.10} \\ \cdashlinelr{1-6}
                                \multirow{10}{*}{\rotatebox[origin=c]{90}{Few-shot}}          & Aya~23~8B                             & 35.73          & 28.37                & 20.65          & 28.25          \\
                                &                                      Aya~23~35B                            & 42.79          & 37.10                & 30.67          & 36.85          \\
                                &                                      Gemma 3 4B                            & 46.72          & 33.37                & 25.07          & 35.05          \\
                                &                                      Gemma 3 12B                          &  54.49          & 40.17                & 33.41          & 42.69          \\
                                &                                      Gemma 3 27B                           & 53.27          & \textbf{46.91}       & \textbf{39.35} & \textbf{46.51} \\
                                &                                      LLaMA~3.1~8B                          & 29.80          & 17.02                & 14.20          & 20.34          \\
                                &                                      LLaMA~3.1~70B                         & 48.23          & 40.99                & 33.37          & 40.86          \\
                                &                                      LLaMA~3.3~70B                         & \textbf{55.02} & 44.36                & 33.76          & 44.38          \\
                                &                                      Orca~2~13B                            & 40.56          & 28.83                & 21.41          & 30.27          \\
                                &                                      GPT-4o mini                           & 54.76          & 40.68                & 33.05          & 42.83          \\ \cdashlinelr{1-6}
                                \multirow{6}{*}{\rotatebox[origin=c]{90}{Fine-tuning}}             & mT5                                   & \textbf{70.74} & \textbf{64.09}       & \textbf{58.06} & \textbf{64.30} \\
                                &                                      Aya~23~8B                             & 69.87          & 61.85                & 57.46          & 63.06          \\
                                &                                      Gemma 3 4B                            & 58.18          & 48.90                & 44.54          & 50.54          \\
                                &                                      Gemma 3 12B                           & 65.23          & 58.56                & 48.45          & 57.41          \\
                                &                                      LLaMA~3.1~8B                          & 66.70          & 60.25                & 56.18          & 61.04          \\
                                &                                      Orca~2~13B                            & 67.13          & 57.93                & 53.85          & 59.64          \\ \bottomrule
\end{tabular}
\end{adjustbox}
\caption{F1 scores in monolingual settings for different models and tasks, alongside with an average result across tasks. \textbf{Bold} results indicate the best result for each combination of \quotes{Type} and task.}
\label{tab:monoresults}
\end{table}

This section presents the results. Among the three tasks, ASTE consistently achieves the highest scores, followed by ASQP, with ACOS being the most challenging. This pattern reflects the increasing complexity of extracting fine-grained sentiment information.

Table~\ref{tab:monoresults} shows the monolingual results. GPT-4o mini achieves the strongest zero-shot results, followed by Gemma~3~27B and LLaMA~3.3~70B, with the latter substantially outperforming its predecessor LLaMA~3.1~70B. Larger and more recent models consistently outperform smaller or older ones. Few-shot examples generally improve performance, though GPT-4o mini shows a slight decrease. Gemma~3~27B achieves the best few-shot results, while Orca~2~13B benefits most from demonstrations, surpassing Aya~23~8B and LLaMA~3.1~8B despite being English-centric. Aya~23~8B consistently outperforms LLaMA~3.1~8B, likely due to its official multilingual support, which includes Czech. Fine-tuned models outperform all zero- and few-shot LLMs, with mT5 achieving the best overall results. Fine-tuned mT5 surpasses the strongest few-shot LLMs by 15--25\% across tasks while offering efficiency advantages in memory and inference speed.

\begin{table}[ht!]
\centering
\begin{adjustbox}{width=0.85\linewidth}
\begin{tabular}{@{}llrrrr@{}}
\toprule
\textbf{Type}                                          & \textbf{Model}                        & \textbf{ASTE}  & \textbf{ASQP}        & \textbf{ACOS}  & \textbf{AVG}   \\ \midrule
 \multirow{6}{*}{\rotatebox[origin=c]{90}{Original}}       & mT5                                   & 49.60          & 41.68                & 35.17          & 42.15          \\
                                &                                      Aya~23~8B                             & 51.02          & 42.12                & 36.41          & 43.18          \\
                                &                                      Gemma 3 4B                            & 49.08          & 36.70                & 30.80          & 38.86          \\
                                &                                      Gemma 3 12B                           & \textbf{53.40} & \textbf{44.17}       & 37.52          & \textbf{45.03} \\
                                &                                      LLaMA~3.1~8B                          & 47.29          & 43.35                & 36.56          & 42.40          \\
                                &                                      Orca~2~13B                            & 48.09          & 41.90                & \textbf{38.32} & 42.77          \\ \cdashlinelr{1-6}
                                 \multirow{6}{*}{\rotatebox[origin=c]{90}{Translated}} & mT5                                   & 51.00          & 40.07                & 35.85          & 42.31          \\
                                &                                      Aya~23~8B                             & 51.59          & 40.56                & 40.46          & 44.20          \\
                                &                                      Gemma 3 4B                            & 48.63          & 39.32                & 32.82          & 40.26          \\
                                &                                      Gemma 3 12B                           & \textbf{57.98} & \textbf{49.13}       & \textbf{41.11} & \textbf{49.41} \\
                                &                                      LLaMA~3.1~8B                          & 49.20          & 44.68                & 37.91          & 43.93          \\
                                &                                      Orca~2~13B                            & 49.29          & 36.62                & 38.68          & 41.53          \\ \bottomrule
\end{tabular}
\end{adjustbox}
\caption{F1 scores in cross-lingual settings for different models and tasks, alongside with an average result across tasks. \textbf{Bold} results indicate the best result for each combination of \quotes{Type} and task.}
\label{tab:crossresults}
\end{table}

Table~\ref{tab:crossresults} shows the cross-lingual results. Training on English only (\quotes{Original}) yields similar results across models, except for Gemma~3~4B, which lags behind. Gemma~3~12B achieves the highest average scores, while Orca~2~13B performs best on ACOS. Adding translated English-to-Czech data (\quotes{Translated}) improves several models: Aya~23~8B and LLaMA~3.1~8B gain over 1\%, while Gemma~3 improves by over 2\% (4B) and 4\% (12B). Gemma~3~12B achieves the best overall cross-lingual results. Nevertheless, performance remains 15--20\% lower than fine-tuned monolingual results, underscoring the challenges of language transfer and annotation inconsistencies. A significant factor is the treatment of opinion terms: English datasets exclude modifiers, whereas Czech datasets include them, resulting in models trained on English data missing modifiers. Cross-lingual performance, however, often surpasses zero- and few-shot monolingual baselines -- for example, Orca~2~13B improves by 30\% over its zero-shot and 12\% over its few-shot performance.

\begin{table}[ht!]
\centering
\begin{adjustbox}{width=0.85\linewidth}
\begin{tabular}{@{}lllrrrr@{}}
\toprule
\textbf{Model}                        & \textbf{ASTE}  & \textbf{ASQP}        & \textbf{ACOS}  & \textbf{AVG}   \\ \midrule

 mT5                                   & \textbf{70.93} & \textbf{64.64}       & \textbf{59.12} & \textbf{64.90}          \\
Aya~23~8B                             & 69.28          & 63.71                & 56.15          & 63.05          \\
Gemma 3 4B                            & 66.01          & 51.44                & 53.90          & 57.11          \\
Gemma 3 12B                           & 68.16          & 61.61 & 56.74          &  62.17         \\
LLaMA~3.1~8B                          & 66.70          & 63.60                & 57.37          & 62.56          \\
Orca~2~13B                            & 67.35          & 57.67                & 53.20          & 59.41          \\ \bottomrule
\end{tabular}
\end{adjustbox}
\caption{F1 scores in multilingual settings for different models and tasks, alongside with an average result across tasks. \textbf{Bold} results indicate the best result for each task.}
\label{tab:multiresults}
\end{table}

Finally, Table~\ref{tab:multiresults} presents results of multilingual experiments, where fine-tuned mT5 again achieves the best overall results, followed by Aya~23~8B, LLaMA~3.1~8B, and Gemma~3~13B, with Gemma~3~4B performing worst. Multilingual fine-tuning generally outperforms cross-lingual setups, as models benefit from training data in both languages, but brings little advantage over monolingual fine-tuning.

Overall, the results highlight trade-offs between fine-tuned models and LLMs. Fine-tuning consistently delivers the best performance, with mT5 combining accuracy with memory and inference efficiency. LLMs remain attractive for rapid deployment, as few-shot prompting and multilingual pre-training boost their performance; however, results vary substantially across models. Larger, newer, and multilingual-aware LLMs perform best. Cross-lingual transfer benefits modestly from machine translation, though dataset inconsistencies limit gains, and multilingual setups do not always surpass monolingual ones. Model choice should therefore balance task complexity, data availability, and computational resources, with fine-tuning as the most reliable strategy and LLMs offering flexible multilingual alternatives.

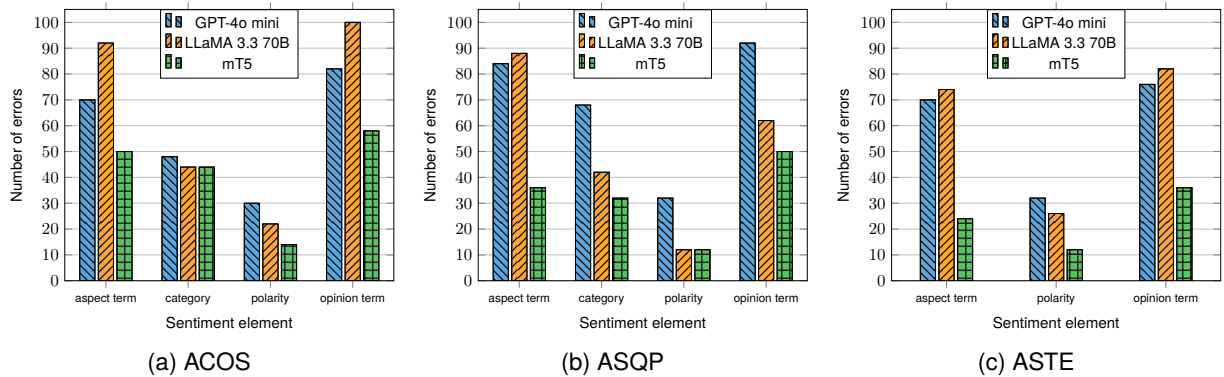
\begin{figure*}[ht!]
    \begin{subfigure}{0.32\textwidth}
        \centering
        \begin{adjustbox}{width=\linewidth}
            \begin{tikzpicture}
                \begin{axis}[
                    ybar,
                    bar width=9pt,
                    xtick={0,1,2,3},
                    ytick={0,10,20,30,40,50,60,70,80,90,100},
                    xticklabels={\scriptsize aspect term,\scriptsize category,\scriptsize polarity,\scriptsize opinion term},
                    ymin=0,
                    ymax=105,
                    xmin=-0.5,
                    xmax=3.5,
                    ymajorgrids=true,
                    ylabel={\footnotesize Number of errors},
                    xlabel={\footnotesize Sentiment element},
                    legend style={at={(0.5,1)}, anchor=north, font=\footnotesize},
                    ]
                    \addplot[black,fill=lightblue,postaction={pattern=north west lines}] coordinates {(0,70) (1,48) (2,30) (3,82)};
                    \addplot[black,fill=lightorange,postaction={pattern=north east lines}] coordinates {(0,92) (1,44) (2,22) (3,100)};
                    \addplot[black,fill=lightgreen,postaction={pattern=grid}] coordinates {(0,50) (1,44) (2,14) (3,58)};
                    \legend{GPT-4o mini, LLaMA~3.3~70B, mT5}
                \end{axis}
            \end{tikzpicture}
        \end{adjustbox}
       \subcaption{ACOS}
    \end{subfigure}
    \hfill
    \begin{subfigure}{0.32\textwidth}
        \centering
        \begin{adjustbox}{width=\linewidth}

            \begin{tikzpicture}
                \begin{axis}[
                    ybar,
                    bar width=9pt,
                    xtick={0,1,2,3},
                    ytick={0,10,20,30,40,50,60,70,80,90,100},
                    xticklabels={\scriptsize aspect term,\scriptsize category,\scriptsize polarity,\scriptsize opinion term},
                    ymin=0,
                    ymax=105,
                    xmin=-0.5,
                    xmax=3.5,
                    ymajorgrids=true,
                    ylabel={\footnotesize Number of errors},
                    xlabel={\footnotesize Sentiment element},
                    legend style={at={(0.5,1)}, anchor=north, font=\footnotesize},
                    ]
                    \addplot[black,fill=lightblue,postaction={pattern=north west lines}] coordinates {(0,84) (1,68) (2,32) (3,92)};
                    \addplot[black,fill=lightorange,postaction={pattern=north east lines}] coordinates {(0,88) (1,42) (2,12) (3,62)};
                    \addplot[black,fill=lightgreen,postaction={pattern=grid}] coordinates {(0,36) (1,32) (2,12) (3,50)};
                    \legend{GPT-4o mini, LLaMA~3.3~70B, mT5}
                \end{axis}
            \end{tikzpicture}
        \end{adjustbox}
       \subcaption{ASQP}
    \end{subfigure}
    \hfill
    \begin{subfigure}{0.32\textwidth}
        \centering
        \begin{adjustbox}{width=\linewidth}
            \begin{tikzpicture}
                \begin{axis}[
                    ybar,
                    bar width=9pt,
                    xtick={0,1,2},
                    ytick={0,10,20,30,40,50,60,70,80,90,100},
                    xticklabels={\scriptsize aspect term,\scriptsize polarity,\scriptsize opinion term},
                    ymin=0,
                    ymax=105,
                    xmin=-0.5,
                    xmax=2.5,
                    ymajorgrids=true,
                    ylabel={\footnotesize Number of errors},
                    xlabel={\footnotesize Sentiment element},
                    legend style={at={(0.5,1)}, anchor=north, font=\footnotesize},
                    ]
                    \addplot[black,fill=lightblue,postaction={pattern=north west lines}] coordinates {(0,70) (1,32) (2,76)};
                    \addplot[black,fill=lightorange,postaction={pattern=north east lines}] coordinates {(0,74) (1,26) (2,82)};
                    \addplot[black,fill=lightgreen,postaction={pattern=grid}] coordinates {(0,24) (1,12) (2,36)};
                    \legend{GPT-4o mini, LLaMA~3.3~70B, mT5}
                \end{axis}
            \end{tikzpicture}
        \end{adjustbox}
        \subcaption{ASTE}
    \end{subfigure}
    \caption{Number of error types for each dataset for GPT-4o mini in zero-shot settings, LLaMA~3.3~70B in few-shot settings, and fine-tuned mT5.}
    \label{fig:error}
\end{figure*}

\subsection{Error Analysis}
To better understand model challenges on our datasets, we manually analyzed 100 randomly sampled examples per dataset for GPT-4o mini (zero-shot), LLaMA~3.3~70B (few-shot), and fine-tuned mT5. Figure~\ref{fig:error} summarizes the error distribution.

The analysis reveals that opinion terms are the most challenging to predict, followed by aspect terms, while sentiment polarity is the easiest to predict. These conclusions align with findings on English datasets~\citep{zhang-etal-2021-aspect-sentiment, smid-etal-2024-llama}. One reason is that opinion terms often span multiple words, whereas aspect terms are usually single words. Implicit opinion terms are particularly difficult, explaining why ACOS yields more errors than ASQP or ASTE. Additional challenges arise from typos, abbreviations, and slang in aspect and opinion terms. In contrast, aspect categories and sentiment polarities are drawn from fixed label sets, making them comparatively easier to predict.

Fine-tuned mT5 consistently produced fewer errors, especially for aspect and opinion terms. Some predictions by GPT-4o mini and LLaMA~3.3~70B, however, could be considered acceptable under alternative annotation interpretations. For instance, models occasionally generalized multiple foods into a single aspect term (e.g. \textit{\quotes{jídlo}} (\textit{\quotes{food}})), predicted canonical morphological forms (e.g. \textit{\quotes{obsluha}} (\textit{\quotes{service}}) instead of \textit{\quotes{obsluhou}}), included plausible but unannotated aspects such as \textit{price}, or merged multiple opinion terms into one (e.g. \textit{\quotes{milá, příjemná}} (\textit{\quotes{nice, pleasant}})) that annotations separate into multiple tuples. Fine-tuning helps align predictions to dataset-specific conventions.

LLMs also produce genuine errors. GPT-4o mini frequently fails to predict any sentiment tuples, while LLaMA~3.3~70B sometimes misclassifies clearly \textit{\quotes{positive}} sentiments, such as \textit{\quotes{hezké prostředí}} (\textit{\quotes{nice environment}}), as \textit{\quotes{negative}}. Both models struggle with idiomatic expressions. For example, \textit{\quotes{Pivečko jak křen}} (idiomatically expressing positive sentiment toward beer) was correctly segmented into aspect and opinion terms. However, LLaMA~3.3~70B predicted \textit{\quotes{negative}} sentiment and GPT-4o mini predicted \textit{\quotes{neutral}}, whereas mT5 correctly predicted \textit{\quotes{positive}}, likely due to exposure during fine-tuning.

Errors in aspect categories typically involve confusion between semantically related classes, such as \textit{\quotes{restaurant general}} vs. \textit{\quotes{restaurant miscellaneous}} or \textit{\quotes{drinks prices}} vs. \textit{\quotes{restaurant prices}}. Rare categories such as \textit{\quotes{location general}} are also problematic. For sentiment polarity, the main issue is misclassifying \textit{\quotes{neutral}} cases -- often mildly positive or negative -- as either \textit{\quotes{positive}} or \textit{\quotes{negative}}, likely due to their low frequency in the dataset.

Generation errors were absent for fine-tuned mT5, contrasting earlier reports~\citep{zhang-etal-2021-aspect-sentiment}. LLaMA~3.3~70B and GPT-4o mini produced only 1--2 formatting errors each, indicating generally reliable sequence generation.

\section{Conclusion}
This paper introduces a manually annotated Czech restaurant dataset for aspect-based sentiment analysis, enriched with opinion term annotations and designed to support three ABSA tasks of varying granularity. Through monolingual, cross-lingual, and multilingual experiments, we evaluate both fine-tuned Transformer-based models and large language models, highlighting strengths and limitations in low-resource settings. Our findings show that fine-tuned models remain the most reliable choice, while LLMs provide flexible alternatives for rapid adaptation across languages.  

To address the challenges of cross-lingual transfer, we propose a translation and label alignment methodology using LLMs, which improves performance and offers a scalable alternative to manual annotation. Error analysis further reveals persistent difficulties, such as detecting subtle opinion terms, disambiguating fine-grained aspect categories, and handling nuanced sentiment expressions in the Czech language. These insights point to clear directions for future research on more robust modelling of opinion expressions and multilingual ABSA.

\section*{Limitations}
While our newly created Czech ABSA dataset provides high-quality annotations, it is limited to the restaurant domain, which may restrict the generalizability of models trained on it to other domains such as hotels, e-commerce, or healthcare. The dataset also focuses exclusively on Czech, so findings may not directly transfer to other languages or dialects. Additionally, implicit opinion terms remain challenging to annotate, and despite review by multiple experienced annotators, some subjectivity may persist, which could affect model performance on subtle sentiment expressions. Finally, the effectiveness of the cross-lingual approach with translation depends on the LLM used for both translation and label alignment.

\section*{Ethics Statement}
During the creation of the new Czech ABSA dataset, the annotators did not identify any instances of content that could be considered discriminatory or overtly racist. Some reviews contain mild offensive language typical of user-generated text, but no systematic bias or harmful content was present in the dataset itself.

We also note that the models used in this study are pre-trained on large internet corpora. As such, they may exhibit unintended biases related to race, gender, or other sensitive attributes due to the nature of the pre-training data.

\section*{Acknowledgements}
The work of Jakub \v{S}m\'{i}d has been supported by the Grant No. SGS-2025-022 -- New Data Processing Methods in Current Areas of Computer Science.
The work of Pavel Kr\'al has been supported by the project R\&D of Technologies for Advanced Digitalization in the Pilsen Metropolitan Area (DigiTech) No. CZ.02.01.01/00/23\_021/0008436. 
Computational resources were provided by the e-INFRA CZ project (ID:90254), supported by the Ministry of Education, Youth and Sports of the Czech Republic.

\section{Bibliographical References}

\bibliographystyle{lrec2026-natbib}
\bibliography{bibliography}

@ARTICLE{absa,
    author={Zhang, Wenxuan and Li, Xin and Deng, Yang and Bing, Lidong and Lam, Wai},
    journal={ IEEE Transactions on Knowledge \& Data Engineering },
    title={{ A Survey on Aspect-Based Sentiment Analysis: Tasks, Methods, and Challenges }},
    year={2023},
    volume={35},
    number={11},
    ISSN={1558-2191},
    pages={11019-11038},
    doi={10.1109/TKDE.2022.3230975},
    url = {https://doi.ieeecomputersociety.org/10.1109/TKDE.2022.3230975},
    publisher={IEEE Computer Society},
    address={Los Alamitos, CA, USA},
    month=nov
}

@article{SMID2025103073,
title = {Cross-lingual aspect-based sentiment analysis: A survey on tasks, approaches, and challenges},
journal = {Information Fusion},
volume = {120},
pages = {103073},
year = {2025},
issn = {1566-2535},
doi = {10.1016/j.inffus.2025.103073},
url = {https://www.sciencedirect.com/science/article/pii/S1566253525001460},
author = {{\v{S}}m{\'i}d, Jakub  and
      Kral, Pavel}
}

@inproceedings{pontiki-etal-2014-semeval,
    title = "{S}em{E}val-2014 Task 4: Aspect Based Sentiment Analysis",
    author = "Pontiki, Maria  and
      Galanis, Dimitris  and
      Pavlopoulos, John  and
      Papageorgiou, Harris  and
      Androutsopoulos, Ion  and
      Manandhar, Suresh",
    editor = "Nakov, Preslav  and
      Zesch, Torsten",
    booktitle = "Proceedings of the 8th International Workshop on Semantic Evaluation ({S}em{E}val 2014)",
    month = aug,
    year = "2014",
    address = "Dublin, Ireland",
    publisher = "Association for Computational Linguistics",
    url = "https://aclanthology.org/S14-2004",
    doi = "10.3115/v1/S14-2004",
    pages = "27--35",
}

@inproceedings{pontiki-etal-2016-semeval,
    title = "{S}em{E}val-2016 Task 5: Aspect Based Sentiment Analysis",
    author = {Pontiki, Maria  and
      Galanis, Dimitris  and
      Papageorgiou, Haris  and
      Androutsopoulos, Ion  and
      Manandhar, Suresh  and
      AL-Smadi, Mohammad  and
      Al-Ayyoub, Mahmoud  and
      Zhao, Yanyan  and
      Qin, Bing  and
      De Clercq, Orph{\'e}e  and
      Hoste, V{\'e}ronique  and
      Apidianaki, Marianna  and
      Tannier, Xavier  and
      Loukachevitch, Natalia  and
      Kotelnikov, Evgeniy  and
      Bel, Nuria  and
      Jim{\'e}nez-Zafra, Salud Mar{\'\i}a  and
      Eryi{\u{g}}it, G{\"u}l{\c{s}}en},
    editor = "Bethard, Steven  and
      Carpuat, Marine  and
      Cer, Daniel  and
      Jurgens, David  and
      Nakov, Preslav  and
      Zesch, Torsten",
    booktitle = "Proceedings of the 10th International Workshop on Semantic Evaluation ({S}em{E}val-2016)",
    month = jun,
    year = "2016",
    address = "San Diego, California",
    publisher = "Association for Computational Linguistics",
    url = "https://aclanthology.org/S16-1002",
    doi = "10.18653/v1/S16-1002",
    pages = "19--30",
}

@inproceedings{pontiki-etal-2015-semeval,
    title = "{S}em{E}val-2015 Task 12: Aspect Based Sentiment Analysis",
    author = "Pontiki, Maria  and
      Galanis, Dimitris  and
      Papageorgiou, Haris  and
      Manandhar, Suresh  and
      Androutsopoulos, Ion",
    editor = "Nakov, Preslav  and
      Zesch, Torsten  and
      Cer, Daniel  and
      Jurgens, David",
    booktitle = "Proceedings of the 9th International Workshop on Semantic Evaluation ({S}em{E}val 2015)",
    month = jun,
    year = "2015",
    address = "Denver, Colorado",
    publisher = "Association for Computational Linguistics",
    url = "https://aclanthology.org/S15-2082",
    doi = "10.18653/v1/S15-2082",
    pages = "486--495",
}

@article{aste,
    title={Knowing What, How and Why: A Near Complete Solution for Aspect-Based Sentiment Analysis},
    volume={34},
    url={https://ojs.aaai.org/index.php/AAAI/article/view/6383},
    DOI={10.1609/aaai.v34i05.6383},
    abstractNote={&lt;p&gt;Target-based sentiment analysis or aspect-based sentiment analysis (ABSA) refers to addressing various sentiment analysis tasks at a fine-grained level, which includes but is not limited to aspect extraction, aspect sentiment classification, and opinion extraction. There exist many solvers of the above individual subtasks or a combination of two subtasks, and they can work together to tell a complete story, i.e. the discussed aspect, the sentiment on it, and the cause of the sentiment. However, no previous ABSA research tried to provide a complete solution in one shot. In this paper, we introduce a new subtask under ABSA, named aspect sentiment triplet extraction (&lt;strong&gt;ASTE&lt;/strong&gt;). Particularly, a solver of this task needs to extract triplets (What, How, Why) from the inputs, which show WHAT the targeted aspects are, HOW their sentiment polarities are and WHY they have such polarities (i.e. opinion reasons). For instance, one triplet from “Waiters are very friendly and the pasta is simply average” could be (‘Waiters’, positive, ‘friendly’). We propose a two-stage framework to address this task. The first stage predicts what, how and why in a unified model, and then the second stage pairs up the predicted what (how) and why from the first stage to output triplets. In the experiments, our framework has set a benchmark performance in this novel triplet extraction task. Meanwhile, it outperforms a few strong baselines adapted from state-of-the-art related methods.&lt;/p&gt;}, number={05}, journal={Proceedings of the AAAI Conference on Artificial Intelligence},
    author={Peng, Haiyun and Xu, Lu and Bing, Lidong and Huang, Fei and Lu, Wei and Si, Luo},
    year={2020},
    month={Apr.},
    pages={8600-8607} 
}

@inproceedings{cai-etal-2021-aspect,
    title = "Aspect-Category-Opinion-Sentiment Quadruple Extraction with Implicit Aspects and Opinions",
    author = "Cai, Hongjie  and
      Xia, Rui  and
      Yu, Jianfei",
    editor = "Zong, Chengqing  and
      Xia, Fei  and
      Li, Wenjie  and
      Navigli, Roberto",
    booktitle = "Proceedings of the 59th Annual Meeting of the Association for Computational Linguistics and the 11th International Joint Conference on Natural Language Processing (Volume 1: Long Papers)",
    month = aug,
    year = "2021",
    address = "Online",
    publisher = "Association for Computational Linguistics",
    url = "https://aclanthology.org/2021.acl-long.29",
    doi = "10.18653/v1/2021.acl-long.29",
    pages = "340--350",
}

@inproceedings{zhang-etal-2021-aspect-sentiment,
    title = "Aspect Sentiment Quad Prediction as Paraphrase Generation",
    author = "Zhang, Wenxuan  and
      Deng, Yang  and
      Li, Xin  and
      Yuan, Yifei  and
      Bing, Lidong  and
      Lam, Wai",
    editor = "Moens, Marie-Francine  and
      Huang, Xuanjing  and
      Specia, Lucia  and
      Yih, Scott Wen-tau",
    booktitle = "Proceedings of the 2021 Conference on Empirical Methods in Natural Language Processing",
    month = nov,
    year = "2021",
    address = "Online and Punta Cana, Dominican Republic",
    publisher = "Association for Computational Linguistics",
    url = "https://aclanthology.org/2021.emnlp-main.726",
    doi = "10.18653/v1/2021.emnlp-main.726",
    pages = "9209--9219",
}

@inproceedings{wolf-etal-2020-transformers,
    title = "Transformers: State-of-the-Art Natural Language Processing",
    author = "Wolf, Thomas  and
      Debut, Lysandre  and
      Sanh, Victor  and
      Chaumond, Julien  and
      Delangue, Clement  and
      Moi, Anthony  and
      Cistac, Pierric  and
      Rault, Tim  and
      Louf, Remi  and
      Funtowicz, Morgan  and
      Davison, Joe  and
      Shleifer, Sam  and
      von Platen, Patrick  and
      Ma, Clara  and
      Jernite, Yacine  and
      Plu, Julien  and
      Xu, Canwen  and
      Le Scao, Teven  and
      Gugger, Sylvain  and
      Drame, Mariama  and
      Lhoest, Quentin  and
      Rush, Alexander",
    editor = "Liu, Qun  and
      Schlangen, David",
    booktitle = "Proceedings of the 2020 Conference on Empirical Methods in Natural Language Processing: System Demonstrations",
    month = oct,
    year = "2020",
    address = "Online",
    publisher = "Association for Computational Linguistics",
    url = "https://aclanthology.org/2020.emnlp-demos.6",
    doi = "10.18653/v1/2020.emnlp-demos.6",
    pages = "38--45",
}

@misc{loshchilov2017decoupled,
      title={Decoupled Weight Decay Regularization}, 
      author={Ilya Loshchilov and Frank Hutter},
      year={2019},
      eprint={1711.05101},
      archivePrefix={arXiv},
      primaryClass={cs.LG},
      url={https://arxiv.org/abs/1711.05101}, 
}

@inproceedings{smid-etal-2024-czech,
    title = "{C}zech Dataset for Complex Aspect-Based Sentiment Analysis Tasks",
    author = "{\v{S}}m{\'\i}d, Jakub  and
      P{\v{r}}ib{\'a}{\v{n}}, Pavel  and
      Prazak, Ondrej  and
      Kral, Pavel",
    editor = "Calzolari, Nicoletta  and
      Kan, Min-Yen  and
      Hoste, Veronique  and
      Lenci, Alessandro  and
      Sakti, Sakriani  and
      Xue, Nianwen",
    booktitle = "Proceedings of the 2024 Joint International Conference on Computational Linguistics, Language Resources and Evaluation (LREC-COLING 2024)",
    month = may,
    year = "2024",
    address = "Torino, Italia",
    publisher = "ELRA and ICCL",
    url = "https://aclanthology.org/2024.lrec-main.384",
    pages = "4299--4310",
}

@inproceedings{klinger-cimiano-2014-usage,
    title = "The {USAGE} review corpus for fine grained multi lingual opinion analysis",
    author = "Klinger, Roman  and
      Cimiano, Philipp",
    editor = "Calzolari, Nicoletta  and
      Choukri, Khalid  and
      Declerck, Thierry  and
      Loftsson, Hrafn  and
      Maegaard, Bente  and
      Mariani, Joseph  and
      Moreno, Asuncion  and
      Odijk, Jan  and
      Piperidis, Stelios",
    booktitle = "Proceedings of the Ninth International Conference on Language Resources and Evaluation ({LREC}'14)",
    month = may,
    year = "2014",
    address = "Reykjavik, Iceland",
    publisher = "European Language Resources Association (ELRA)",
    url = "http://www.lrec-conf.org/proceedings/lrec2014/pdf/85_Paper.pdf",
    pages = "2211--2218",
}

@inproceedings{saeidi-etal-2016-sentihood,
    title = "{S}enti{H}ood: Targeted Aspect Based Sentiment Analysis Dataset for Urban Neighbourhoods",
    author = "Saeidi, Marzieh  and
      Bouchard, Guillaume  and
      Liakata, Maria  and
      Riedel, Sebastian",
    booktitle = "Proceedings of {COLING} 2016, the 26th International Conference on Computational Linguistics: Technical Papers",
    month = dec,
    year = "2016",
    address = "Osaka, Japan",
    publisher = "The COLING 2016 Organizing Committee",
    url = "https://aclanthology.org/C16-1146",
    pages = "1546--1556",
}

@inproceedings{xu-etal-2020-position,
    title = "Position-Aware Tagging for Aspect Sentiment Triplet Extraction",
    author = "Xu, Lu  and
      Li, Hao  and
      Lu, Wei  and
      Bing, Lidong",
    editor = "Webber, Bonnie  and
      Cohn, Trevor  and
      He, Yulan  and
      Liu, Yang",
    booktitle = "Proceedings of the 2020 Conference on Empirical Methods in Natural Language Processing (EMNLP)",
    month = nov,
    year = "2020",
    address = "Online",
    publisher = "Association for Computational Linguistics",
    url = "https://aclanthology.org/2020.emnlp-main.183",
    doi = "10.18653/v1/2020.emnlp-main.183",
    pages = "2339--2349",
}

@inproceedings{fan-etal-2019-target,
    title = "Target-oriented Opinion Words Extraction with Target-fused Neural Sequence Labeling",
    author = "Fan, Zhifang  and
      Wu, Zhen  and
      Dai, Xin-Yu  and
      Huang, Shujian  and
      Chen, Jiajun",
    editor = "Burstein, Jill  and
      Doran, Christy  and
      Solorio, Thamar",
    booktitle = "Proceedings of the 2019 Conference of the North {A}merican Chapter of the Association for Computational Linguistics: Human Language Technologies, Volume 1 (Long and Short Papers)",
    month = jun,
    year = "2019",
    address = "Minneapolis, Minnesota",
    publisher = "Association for Computational Linguistics",
    url = "https://aclanthology.org/N19-1259",
    doi = "10.18653/v1/N19-1259",
    pages = "2509--2518",
}

@inproceedings{steinberger-etal-2014-aspect,
    title = "Aspect-Level Sentiment Analysis in {C}zech",
    author = "Steinberger, Josef  and
      Brychc{\'\i}n, Tom{\'a}{\v{s}}  and
      Konkol, Michal",
    booktitle = "Proceedings of the 5th Workshop on Computational Approaches to Subjectivity, Sentiment and Social Media Analysis",
    month = jun,
    year = "2014",
    address = "Baltimore, Maryland",
    publisher = "Association for Computational Linguistics",
    url = "https://aclanthology.org/W14-2605",
    doi = "10.3115/v1/W14-2605",
    pages = "24--30",
}

@inproceedings{vaswani2017attention,
 author = {Vaswani, Ashish and Shazeer, Noam and Parmar, Niki and Uszkoreit, Jakob and Jones, Llion and Gomez, Aidan N and Kaiser, \L ukasz and Polosukhin, Illia},
 booktitle = {Advances in Neural Information Processing Systems},
 editor = {I. Guyon and U. Von Luxburg and S. Bengio and H. Wallach and R. Fergus and S. Vishwanathan and R. Garnett},
 publisher = {Curran Associates, Inc.},
 title = {Attention is All you Need},
 url = {https://proceedings.neurips.cc/paper_files/paper/2017/file/3f5ee243547dee91fbd053c1c4a845aa-Paper.pdf},
 volume = {30},
 pages = {6000–6010},
 year = {2017}
}

@inproceedings{priban-prazak-2023-improving,
    title = "Improving Aspect-Based Sentiment with End-to-End Semantic Role Labeling Model",
    author = "P{\v{r}}ib{\'a}{\v{n}}, Pavel  and
      Pra\v{z}{\'a}k, Ond\v{r}ej",
    editor = "Mitkov, Ruslan  and
      Angelova, Galia",
    booktitle = "Proceedings of the 14th International Conference on Recent Advances in Natural Language Processing",
    month = sep,
    year = "2023",
    address = "Varna, Bulgaria",
    publisher = "INCOMA Ltd., Shoumen, Bulgaria",
    url = "https://aclanthology.org/2023.ranlp-1.96",
    pages = "888--897",
}

@inproceedings{SLON-Lenc2016Neural,
	title = {Neural Networks for Sentiment Analysis in Czech},
	author = {Ladislav Lenc and Tom{\'a}s Hercig},
	booktitle = {Proceedings of the 16th {ITAT}: Slovensko{\v{c}}esk{\'{y}} {NLP} workshop (Slo{NLP} 2016)},
	editor = {Bro{\v{n}}a Brejov{\'{a}}},
	year = {2016},
	publisher = {CreateSpace Independent Publishing Platform},
	organization = {Comenius University in Bratislava, Faculty of Mathematics, Physics and Informatics},
	address = {Bratislava, Slovakia},
	venue = {{SOREA} Hutn{\'{i}}k I.},
	series = {{CEUR} Workshop Proceedings},
	volume = {1649},
	pages = {48-55},
	isbn = {978-1537016740},
	issn = {1613-0073}
}

@article{hercig2016unsupervised,
  title={Unsupervised methods to improve aspect-based sentiment analysis in Czech},
  author={Hercig, Tom{\'a}{\v{s}} and Brychc{\'\i}n, Tom{\'a}{\v{s}} and Svoboda, Luk{\'a}{\v{s}} and Konkol, Michal and Steinberger, Josef},
  journal={Computaci{\'o}n y Sistemas},
  volume={20},
  number={3},
  pages={365--375},
  year={2016},
  publisher={Instituto Polit{\'e}cnico Nacional, Centro de Investigaci{\'o}n en Computaci{\'o}n}
}

@inproceedings{tamchyna2015czech,
  title={Czech Aspect-Based Sentiment Analysis: A New Dataset and Preliminary Results.},
  author={Tamchyna, Ales and Fiala, Ondrej and Veselovsk{\'a}, Katerina},
  booktitle={ITAT},
  pages={95--99},
  year={2015}
}

@conference{icaart25,
author={{\v{S}}m{\'i}d, Jakub  and
      P{\v{r}}ib{\'a}{\v{n}}, Pavel  and
      Kral, Pavel},
title={Advancing Cross-Lingual Aspect-Based Sentiment Analysis with LLMs and Constrained Decoding for Sequence-to-Sequence Models},
booktitle={Proceedings of the 17th International Conference on Agents and Artificial Intelligence - Volume 2: ICAART},
year={2025},
pages={757-766},
publisher={SciTePress},
organization={INSTICC},
doi={10.5220/0013349400003890},
isbn={978-989-758-737-5},
}

@article{Wu_2025,
title={Evaluating zero-shot multilingual aspect-based sentiment analysis with large language models},
ISSN={1868-808X},
url={http://dx.doi.org/10.1007/s13042-025-02711-z},
DOI={10.1007/s13042-025-02711-z},
journal={International Journal of Machine Learning and Cybernetics},
publisher={Springer Science and Business Media LLC},
author={Wu, Chengyan and Ma, Bolei and Zhang, Zheyu and Deng, Ningyuan and He, Yanqing and Xue, Yun}, 
year={2025},
month=jun }

@inproceedings{smid-etal-2025-laca,
    title = "{LACA}: Improving Cross-lingual Aspect-Based Sentiment Analysis with {LLM} Data Augmentation",
    author = "{\v{S}}m{\'i}d, Jakub  and
      Priban, Pavel  and
      Kral, Pavel",
    editor = "Che, Wanxiang  and
      Nabende, Joyce  and
      Shutova, Ekaterina  and
      Pilehvar, Mohammad Taher",
    booktitle = "Proceedings of the 63rd Annual Meeting of the Association for Computational Linguistics (Volume 1: Long Papers)",
    month = jul,
    year = "2025",
    address = "Vienna, Austria",
    publisher = "Association for Computational Linguistics",
    url = "https://aclanthology.org/2025.acl-long.41/",
    doi = "10.18653/v1/2025.acl-long.41",
    pages = "839--853",
    ISBN = "979-8-89176-251-0"
}

@inproceedings{smid-etal-2024-llama,
    title = "{LL}a{MA}-Based Models for Aspect-Based Sentiment Analysis",
    author = "{\v{S}}m{\'\i}d, Jakub  and
      Priban, Pavel  and
      Kral, Pavel",
    editor = "De Clercq, Orph{\'e}e  and
      Barriere, Valentin  and
      Barnes, Jeremy  and
      Klinger, Roman  and
      Sedoc, Jo{\~a}o  and
      Tafreshi, Shabnam",
    booktitle = "Proceedings of the 14th Workshop on Computational Approaches to Subjectivity, Sentiment, {\&} Social Media Analysis",
    month = aug,
    year = "2024",
    address = "Bangkok, Thailand",
    publisher = "Association for Computational Linguistics",
    url = "https://aclanthology.org/2024.wassa-1.6",
    doi = "10.18653/v1/2024.wassa-1.6",
    pages = "63--70",
}

@inproceedings{dyer-etal-2013-simple,
    title = "A Simple, Fast, and Effective Reparameterization of {IBM} Model 2",
    author = "Dyer, Chris  and
      Chahuneau, Victor  and
      Smith, Noah A.",
    editor = "Vanderwende, Lucy  and
      Daum{\'e} III, Hal  and
      Kirchhoff, Katrin",
    booktitle = "Proceedings of the 2013 Conference of the North {A}merican Chapter of the Association for Computational Linguistics: Human Language Technologies",
    month = jun,
    year = "2013",
    address = "Atlanta, Georgia",
    publisher = "Association for Computational Linguistics",
    url = "https://aclanthology.org/N13-1073/",
    pages = "644--648"
}

@inproceedings{zhang-etal-2021-cross,
    title = "Cross-lingual Aspect-based Sentiment Analysis with Aspect Term Code-Switching",
    author = "Zhang, Wenxuan  and
      He, Ruidan  and
      Peng, Haiyun  and
      Bing, Lidong  and
      Lam, Wai",
    editor = "Moens, Marie-Francine  and
      Huang, Xuanjing  and
      Specia, Lucia  and
      Yih, Scott Wen-tau",
    booktitle = "Proceedings of the 2021 Conference on Empirical Methods in Natural Language Processing",
    month = nov,
    year = "2021",
    address = "Online and Punta Cana, Dominican Republic",
    publisher = "Association for Computational Linguistics",
    url = "https://aclanthology.org/2021.emnlp-main.727/",
    doi = "10.18653/v1/2021.emnlp-main.727",
    pages = "9220--9230"
}

@misc{gemmateam2025gemma3technicalreport,
      title={Gemma 3 Technical Report}, 
      author={Gemma Team and Aishwarya Kamath and others},
      year={2025},
      eprint={2503.19786},
      archivePrefix={arXiv},
      primaryClass={cs.CL},
      url={https://arxiv.org/abs/2503.19786}, 
}

@InProceedings{smid2025largelanguagemodelsczech,
author="{\v{S}}m{\'i}d, Jakub
and P{\v{r}}ib{\'a}{\v{n}}, Pavel
and Kr{\'a}l, Pavel",
editor="Ek{\v{s}}tein, Kamil
and Konop{\'i}k, Miloslav
and Pra{\v{z}}{\'a}k, Ond{\v{r}}ej
and P{\'a}rtl, Franti{\v{s}}ek",
title="Large Language Models for Czech Aspect-Based Sentiment Analysis",
booktitle="Text, Speech, and Dialogue",
year="2026",
publisher="Springer Nature Switzerland",
address="Cham",
pages="15--26",
isbn="978-3-032-02551-7"
}

@inproceedings{zhang-etal-2024-sentiment,
    title = "Sentiment Analysis in the Era of Large Language Models: A Reality Check",
    author = "Zhang, Wenxuan  and
      Deng, Yue  and
      Liu, Bing  and
      Pan, Sinno  and
      Bing, Lidong",
    editor = "Duh, Kevin  and
      Gomez, Helena  and
      Bethard, Steven",
    booktitle = "Findings of the Association for Computational Linguistics: NAACL 2024",
    month = jun,
    year = "2024",
    address = "Mexico City, Mexico",
    publisher = "Association for Computational Linguistics",
    url = "https://aclanthology.org/2024.findings-naacl.246",
    doi = "10.18653/v1/2024.findings-naacl.246",
    pages = "3881--3906",
}

@inproceedings{xianlong-etal-2023-tagging,
    title = "Tagging-Assisted Generation Model with Encoder and Decoder Supervision for Aspect Sentiment Triplet Extraction",
    author = "Xianlong, Luo  and
      Yang, Meng  and
      Wang, Yihao",
    editor = "Bouamor, Houda  and
      Pino, Juan  and
      Bali, Kalika",
    booktitle = "Proceedings of the 2023 Conference on Empirical Methods in Natural Language Processing",
    month = dec,
    year = "2023",
    address = "Singapore",
    publisher = "Association for Computational Linguistics",
    url = "https://aclanthology.org/2023.emnlp-main.129",
    doi = "10.18653/v1/2023.emnlp-main.129",
    pages = "2078--2093",
}

@inproceedings{gao-etal-2022-lego,
    title = "{LEGO}-{ABSA}: A Prompt-based Task Assemblable Unified Generative Framework for Multi-task Aspect-based Sentiment Analysis",
    author = "Gao, Tianhao  and
      Fang, Jun  and
      Liu, Hanyu  and
      Liu, Zhiyuan  and
      Liu, Chao  and
      Liu, Pengzhang  and
      Bao, Yongjun  and
      Yan, Weipeng",
    editor = "Calzolari, Nicoletta  and
      Huang, Chu-Ren  and
      Kim, Hansaem  and
      Pustejovsky, James  and
      Wanner, Leo  and
      Choi, Key-Sun  and
      Ryu, Pum-Mo  and
      Chen, Hsin-Hsi  and
      Donatelli, Lucia  and
      Ji, Heng  and
      Kurohashi, Sadao  and
      Paggio, Patrizia  and
      Xue, Nianwen  and
      Kim, Seokhwan  and
      Hahm, Younggyun  and
      He, Zhong  and
      Lee, Tony Kyungil  and
      Santus, Enrico  and
      Bond, Francis  and
      Na, Seung-Hoon",
    booktitle = "Proceedings of the 29th International Conference on Computational Linguistics",
    month = oct,
    year = "2022",
    address = "Gyeongju, Republic of Korea",
    publisher = "International Committee on Computational Linguistics",
    url = "https://aclanthology.org/2022.coling-1.610",
    pages = "7002--7012",
}

@inproceedings{mao-etal-2022-seq2path,
    title = "{S}eq2{P}ath: Generating Sentiment Tuples as Paths of a Tree",
    author = "Mao, Yue  and
      Shen, Yi  and
      Yang, Jingchao  and
      Zhu, Xiaoying  and
      Cai, Longjun",
    editor = "Muresan, Smaranda  and
      Nakov, Preslav  and
      Villavicencio, Aline",
    booktitle = "Findings of the Association for Computational Linguistics: ACL 2022",
    month = may,
    year = "2022",
    address = "Dublin, Ireland",
    publisher = "Association for Computational Linguistics",
    url = "https://aclanthology.org/2022.findings-acl.174",
    doi = "10.18653/v1/2022.findings-acl.174",
    pages = "2215--2225",
}

@inproceedings{smid-priban-2023-prompt,
    title = "Prompt-Based Approach for {C}zech Sentiment Analysis",
    author = "{\v{S}}m{\'\i}d, Jakub  and
      P{\v{r}}ib{\'a}{\v{n}}, Pavel",
    editor = "Mitkov, Ruslan  and
      Angelova, Galia",
    booktitle = "Proceedings of the 14th International Conference on Recent Advances in Natural Language Processing",
    month = sep,
    year = "2023",
    address = "Varna, Bulgaria",
    publisher = "INCOMA Ltd., Shoumen, Bulgaria",
    url = "https://aclanthology.org/2023.ranlp-1.118",
    pages = "1110--1120",
}

@inproceedings{zhang-etal-2021-towards-generative,
    title = "Towards Generative Aspect-Based Sentiment Analysis",
    author = "Zhang, Wenxuan  and
      Li, Xin  and
      Deng, Yang  and
      Bing, Lidong  and
      Lam, Wai",
    booktitle = "Proceedings of the 59th Annual Meeting of the Association for Computational Linguistics and the 11th International Joint Conference on Natural Language Processing (Volume 2: Short Papers)",
    month = aug,
    year = "2021",
    address = "Online",
    publisher = "Association for Computational Linguistics",
    url = "https://aclanthology.org/2021.acl-short.64",
    doi = "10.18653/v1/2021.acl-short.64",
    pages = "504--510",
}

@inproceedings{gou-etal-2023-mvp,
    title = "{M}v{P}: Multi-view Prompting Improves Aspect Sentiment Tuple Prediction",
    author = "Gou, Zhibin  and
      Guo, Qingyan  and
      Yang, Yujiu",
    booktitle = "Proceedings of the 61st Annual Meeting of the Association for Computational Linguistics (Volume 1: Long Papers)",
    month = jul,
    year = "2023",
    address = "Toronto, Canada",
    publisher = "Association for Computational Linguistics",
    url = "https://aclanthology.org/2023.acl-long.240",
    doi = "10.18653/v1/2023.acl-long.240",
    pages = "4380--4397",
}

@inproceedings{xue-etal-2021-mt5,
    title = "m{T}5: A Massively Multilingual Pre-trained Text-to-Text Transformer",
    author = "Xue, Linting  and
      Constant, Noah  and
      Roberts, Adam  and
      Kale, Mihir  and
      Al-Rfou, Rami  and
      Siddhant, Aditya  and
      Barua, Aditya  and
      Raffel, Colin",
    editor = "Toutanova, Kristina  and
      Rumshisky, Anna  and
      Zettlemoyer, Luke  and
      Hakkani-Tur, Dilek  and
      Beltagy, Iz  and
      Bethard, Steven  and
      Cotterell, Ryan  and
      Chakraborty, Tanmoy  and
      Zhou, Yichao",
    booktitle = "Proceedings of the 2021 Conference of the North American Chapter of the Association for Computational Linguistics: Human Language Technologies",
    month = jun,
    year = "2021",
    address = "Online",
    publisher = "Association for Computational Linguistics",
    url = "https://aclanthology.org/2021.naacl-main.41",
    doi = "10.18653/v1/2021.naacl-main.41",
    pages = "483--498",
}

@inproceedings{qlora,
 author = {Dettmers, Tim and Pagnoni, Artidoro and Holtzman, Ari and Zettlemoyer, Luke},
 booktitle = {Advances in Neural Information Processing Systems},
 editor = {A. Oh and T. Naumann and A. Globerson and K. Saenko and M. Hardt and S. Levine},
 pages = {10088--10115},
 publisher = {Curran Associates, Inc.},
 title = {QLoRA: Efficient Finetuning of Quantized LLMs},
 url = {https://proceedings.neurips.cc/paper_files/paper/2023/file/1feb87871436031bdc0f2beaa62a049b-Paper-Conference.pdf},
 volume = {36},
 year = {2023}
}

@inproceedings{hu2022lora,
    title={Lo{RA}: Low-Rank Adaptation of Large Language Models},
    author={Edward J Hu and Yelong Shen and Phillip Wallis and Zeyuan Allen-Zhu and Yuanzhi Li and Shean Wang and Lu Wang and Weizhu Chen},
    booktitle={International Conference on Learning Representations},
    year={2022},
    url={https://openreview.net/forum?id=nZeVKeeFYf9}
}

@manual{openai2024gpt4o,
  title        = {GPT-4o},
  author       = {OpenAI},
  year         = {2024},
  url          = {https://platform.openai.com/docs/models/o1},
  note         = {Accessed November 2024},
}

@misc{dubey2024llama3herdmodels,
      title={The Llama 3 Herd of Models}, 
      author={Abhimanyu Dubey and Abhinav Jauhri and Abhinav Pandey and others},
      year={2024},
      eprint={2407.21783},
      archivePrefix={arXiv},
      primaryClass={cs.AI},
      url={https://arxiv.org/abs/2407.21783}, 
}

@misc{mitra2023orca,
      title={Orca 2: Teaching Small Language Models How to Reason}, 
      author={Arindam Mitra and Luciano Del Corro and Shweti Mahajan and Andres Codas and Clarisse Simoes and Sahaj Agarwal and Xuxi Chen and Anastasia Razdaibiedina and Erik Jones and Kriti Aggarwal and Hamid Palangi and Guoqing Zheng and Corby Rosset and Hamed Khanpour and Ahmed Awadallah},
      year={2023},
      eprint={2311.11045},
      archivePrefix={arXiv},
      primaryClass={cs.AI},
      url={https://arxiv.org/abs/2311.11045}, 
}

@misc{aryabumi2024aya23openweight,
      title={Aya 23: Open Weight Releases to Further Multilingual Progress}, 
      author={Viraat Aryabumi and John Dang and Dwarak Talupuru and Saurabh Dash and David Cairuz and Hangyu Lin and Bharat Venkitesh and Madeline Smith and Jon Ander Campos and Yi Chern Tan and Kelly Marchisio and Max Bartolo and Sebastian Ruder and Acyr Locatelli and Julia Kreutzer and Nick Frosst and Aidan Gomez and Phil Blunsom and Marzieh Fadaee and Ahmet Üstün and Sara Hooker},
      year={2024},
      eprint={2405.15032},
      archivePrefix={arXiv},
      primaryClass={cs.CL},
      url={https://arxiv.org/abs/2405.15032}, 
}

\end{document}